%% file: root.tex
\documentclass[letterpaper, 10 pt, conference]{ieeeconf}  

\IEEEoverridecommandlockouts                              

\usepackage{graphics} 
\usepackage{amsmath} 
\usepackage{amssymb}  
\usepackage{booktabs} 
\usepackage{algorithm} 
\usepackage{algpseudocode} 
\usepackage{tabularx} 
\usepackage{tikz} 
\usepackage{url}
\usepackage{todonotes}

\title{\LARGE \bf
Bifurcation Identification
for Ultrasound-driven Robotic Cannulation
}

\author{Cecilia G. Morales$^{1}$, Dhruv Srikanth$^{1}$, Jack H. Good$^{1}$, Keith A. Dufendach$^{2}$, Artur Dubrawski$^{1}$ \\
\thanks{This work was partially supported by the U.S.\ Army award W81XWH-19-C0083. The authors would like to thank Nico Zevallos, Dr.\ Michael R.\ Pinsky, Dr.\ Hernando Gomez. and Dr.\ Leonard Weiss for gathering and adjudicating the data for our experiments. Special thanks to Joseph Merante for his assistance with the figures. Lastly, we would like to thank Mononito Goswami for insightful comments and proofreading this manuscript.}
\thanks{$^{1}$ Authors are with Carnegie Mellon University, Pittsburgh, PA, USA
{\tt\small \{cgmorale, dsrikant, jhgood, awd\}@andrew.cmu.edu}}
\thanks{$^{2}$ Author is a surgeon with the Department of Cardiothoracic Surgery, University of Pittsburgh School of Medicine, Pittsburgh, PA, USA 
{\tt\small dufendachka@upmc.edu}}}
\begin{document}

\maketitle
\thispagestyle{empty}
\pagestyle{empty}

\begin{abstract}

In trauma and critical care settings, rapid and precise intravascular access is key to patients' survival. Our research aims at ensuring this access, even when skilled medical personnel are not readily available. Vessel bifurcations are anatomical landmarks that can guide the safe placement of catheters or needles during medical procedures. 
Although ultrasound is advantageous in navigating anatomical landmarks in emergency scenarios due to its portability and safety, to our knowledge no existing algorithm can autonomously extract vessel bifurcations using ultrasound images. This is primarily due to the limited availability of ground truth data, in particular, data from live subjects, needed for training and validating reliable models. 
We introduce BIFURC (\underline{B}ifurcation \underline{I}dentification \underline{F}or \underline{U}ltrasound-driven \underline{R}obot \underline{C}annulation), a novel algorithm that identifies vessel bifurcations and provides optimal needle insertion sites for an autonomous robotic cannulation system. BIFURC integrates expert knowledge with deep learning techniques to efficiently detect vessel bifurcations within the femoral region and can be trained on a limited amount of in-vivo data. We evaluated our algorithm using a medical phantom as well as real-world experiments involving live pigs. In all cases, BIFURC consistently identified bifurcation points and needle insertion locations in alignment with those identified by expert clinicians. 
\end{abstract}

\section{INTRODUCTION}
\label{sec:introduction}

\input{Sections/introduction}

\section{Background and Related Work}
\label{sec:relatedWorks}
\input{Sections/relatedworksReal}

\section{METHODS}
\label{sec:method}
\input{Sections/method}

\section{EXPERIMENTS}
\label{sec:experiments}
\input{Sections/Experiments}

\section{RESULTS AND ANALYSIS}
\label{sec:results}

\input{Sections/Results}

\section{CONCLUSION}
\label{sec:conclusion}
\input{Sections/conclusion}






\bibliographystyle{IEEEtran}
\bibliography{bibliography}

\addtolength{\textheight}{-12cm}


\end{document}

%% file: Sections/introduction.tex
In trauma and critical care, adequate arterial and venous access can mean the difference between life and death~\cite{psnet_cvc_placement}. Catheters allow for rapid administration of medications and fluids, live hemodynamic monitoring, and occasionally life support options through extracorporeal membrane oxygenation (ECMO) or resuscitative endovascular balloon occlusion of the aorta (REBOA). These options are essential for the optimal care of accident victims or individuals wounded in underserved areas or in combat scenarios~\cite{nowadly2020use}. Appropriate insertion of needles and catheters into central arteries and veins requires expert medical personnel, often working under adverse circumstances and time pressure. Since adequate access to experienced staff is not always possible, especially in mass casualty events, we develop an automated system that enables novice medical providers with limited training to initiate intravascular treatment safely and quickly. 

\begin{figure}[tp]
\centering\includegraphics[width=1.0\columnwidth]{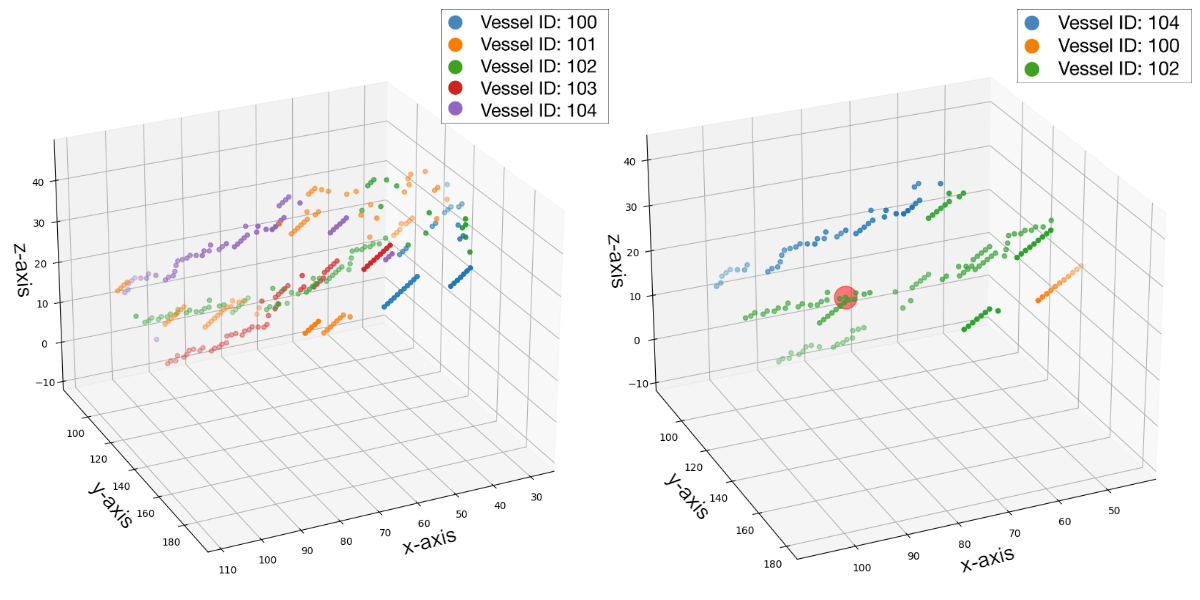}
  \caption{Left: Center lines of vessels in a real-world experiment. Without the BIFURC algorithm, the presence of noise greatly hampers the identification of individual vessels and subsequent detection of bifurcations. In contrast, data processed with BIFURC (right) yields a clearer depiction of distinct vessels and the relevant bifurcation, denoted by the red circle. Vessel IDs and colors have been chosen arbitrarily.}
  \label{effect_of_merging}
  \vspace{-10pt}
\end{figure}

Identifying the patient's anatomy is crucial for gaining intravascular access. According to Rupp's rule, the bifurcation of the common femoral artery into the superficial femoral artery and profunda femoris artery is a key anatomical landmark for safely performing femoral arterial and venous cannulation~\cite{Bangalore2011}.

Given the importance of precise anatomical identification, ultrasound (US) imaging emerges as the optimal tool for navigating intravascular access in emergency and field care scenarios~\cite{gjesteby2022ultrasound}. Its portability, versatility, affordability and accessibility surpass other modalities such as X-ray, computed tomography (CT), or magnetic resonance imaging (MRI)~\cite{gjesteby2022ultrasound}. 



Despite its potential, US-guided central access is rarely utilized in prehospital settings. Additionally, there has been limited effort to integrate autonomous robotic ultrasound scanning into minimally invasive procedures~\cite{Marahrens2022TowardsAR}. The key barrier is the lack of advanced US interpretation skills and expertise in image-guided needle placement among emergency medicine field personnel, exacerbating the risk of complications~\cite{Tse2022Dec}. 

We automatically detect vessel bifurcations using a robotic ultrasound system. 
Below, we summarize the key contributions of our work:

\begin{enumerate}
    \item 
    To accurately and autonomously identify vessel bifurcations, we propose the first method to create a 3D vessel skeletonization using a linear ultrasound probe. 
    \item The first algorithm to automatically detect an optimal needle insertion point in the femoral area.
    \item Evaluation on a greater number of real-world instances vs.\ any other machine learning study of needle insertion in the femoral vessel using ultrasound.
    \item Notably, our algorithm is the first to achieve expert-level performance by autonomously identifying bifurcations and accurately determining the needle insertion site in both simulated data and in live pigs. 
\end{enumerate}

%% file: Sections/relatedworksReal.tex
\begin{figure*}
  \centering
  \def\svgwidth{\textwidth}
  \includegraphics[width=1.0\textwidth]{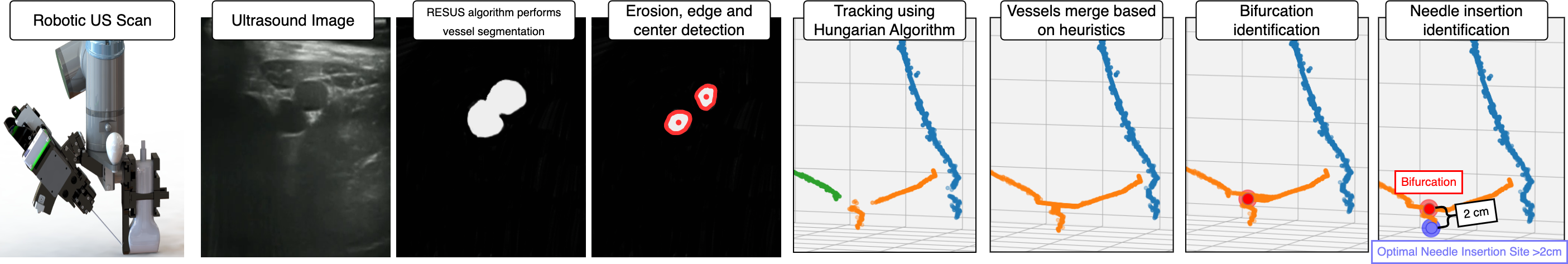}
  \vspace{-10pt}
  \caption{BIFURC is a deep learning technique augmented with expert-derived heuristics designed to identify bifurcations and optimal needle insertion sites. First, the robot scans the leg and collects 2D ultrasound images alongside their poses. We then utilize a model to segment the vessels from these images. Next, we apply an erosion algorithm to distinguish vessels with overlapping segmentation masks. 
  Using robot poses and vessel centerlines, we apply heuristics to track and merge distinct segments to identify vessel bifurcations. Finally, we locate a \textit{safe} needle insertion spot, which is at least 2cm away from the identified bifurcation.}
  \label{fig:methods}
  \vspace{-15pt}
\end{figure*}

Although research on autonomous needle insertion with ultrasound in femoral vessels is limited~\cite{Koskinipolou}, some advancements have been made in related areas, such as femoral vessel visualization techniques~\cite{Marahrens2022TowardsAR, Morales2023VizTool, Smistad2016}, segmentation uncertainty~\cite{asmus2024}, vessel deformability~\cite{Bal2023}, and needle tracking~\cite{goel2024motion,Scholten2017}. Despite those developments, identifying the optimal needle insertion site remains an underexplored and novel area of investigation. We draw insights from three review papers on medical imaging and vessel segmentation~\cite{Brattain2018Apr, Moccia2018May, Kirbas2004Jun}. 
Our approach is inspired by the performance boost observed when deep learning techniques are integrated with manually crafted features, rather than relying solely on one methodology.

\subsubsection{Autonomous and Semi-Autonomous Robotic Systems for Optimal Needle Insertion Location}
Chen et al. presented a benchtop system combining infrared and ultrasound technologies alongside deep learning algorithms to aid in needle selection and insertion~\cite{Chen2020Feb}. However, unlike our approach, their system predominantly targeted peripheral vessels near the skin's surface. This is also seen in \cite{9484669}. Consequently, they leveraged infrared imaging to visualize the vessel structure beneath the skin but lacked the ability to address larger, deeper vessels essential for critical medical procedures. This limitation was similarly noted in Cheng et al.'s Cathbot proposal, which required practitioner intervention and encountered comparable restrictions~\cite{Cheng2019Jan}.

In contrast, AI-GUIDE, a semi-autonomous device aimed at assisting users with femoral US-guided vascular catheterization~\cite{bios11120522}, showed promising outcomes on three pigs. However, AI-GUIDE relies primarily on classification techniques to detect vessel bifurcations, which does not utilize temporal data. In our preliminary tests, we encountered limitations with the multiclass classification YOLO model employed by AI-GUIDE, yielding an average intersection over union (IoU) in vessel detection of 0.23, as seen in Table~\ref{tab:resus-vs-yolo}, which is insufficient for vessel identification purposes.
Instead, our approach relies on segmentation and reconstruction, which enhance its ability to accurately identify bifurcations. It is important to note that AI-GUIDE's reliance on human intervention introduces the potential for errors, a factor that our fully automated system seeks to address. 



\subsubsection{Vessel Extraction}

Vessel extraction involves outlining blood vessels in medical images for analysis. Although research has largely focused on stationary imaging modalities such as CT, MRI, and X-rays~\cite{Kirbas2004Jun}, automated US segmentation faces distinct challenges. US images are frequently affected by issues like speckle noise, shadowing, and incomplete boundaries~\cite{Brattain2018Apr}, which can obscure vessel outlines and hinder delineation of vessel boundaries and centerlines, as shown in Fig.~\ref{resus}. Moreover, although some methods for extracting vessels using ultrasound exist as in~\cite{Marahrens2022TowardsAR}, they are typically limited to phantoms and do not effectively address more advanced capabilities such as autonomous identification of bifurcations, as shown in~\cite{Smistad2016}. 

\subsubsection{Vessel Tracking}
 Vessel tracking refers to the process of locating and monitoring the movement or trajectory of vessels within the human body. Vessel tracking is important for medical procedures including catheterization \cite{Moccia2018May, Kirbas2004Jun, Wang2009Jun}. Several algorithms have explored Kalman filters (KFs) to refine tracking accuracy \cite{Smistad2016,10.1007/978-3-030-00937-3_85, TAGHAVI2022106695}. However, KFs introduce spurious data points which can hinder downstream tasks, such as bifurcation identification and optimal needle insertion point detection. Some other studies model 3D geometry information using Bayesian 3D U-Net models. But these models are computationally expensive and have lower IoU scores in comparison to 2D techniques (cf.\ Tab.~\ref{tab:resus-vs-yolo}). Recognizing the computational hurdles with the implementation of deep learning 3D models, particularly in embedded systems, underscores the need for innovative approaches~\cite{Tetteh2020Dec}. We propose training 2D deep learning models while utilizing 3D structural information captured at multiple robot poses. This approach aims to mitigate computational burdens and improve the performance of portable, fully autonomous, and quasi-real-time systems for vessel tracking.

\subsubsection{Spatio-Temporal Information for Vessel Bifurcation Identification}
The US technology presents a significant advantage over other imaging modalities due to its capacity to generate real-time video. Within fields such as echocardiography and obstetrics, the integration of machine learning techniques to support interpretation of US data has become increasingly prevalent, leveraging spatio-temporal data to enhance diagnostic accuracy and treatment outcomes~\cite{Brattain2018Apr}. While video clips offer a richer source of information compared to individual image frames, there is a notable gap in research regarding their application in identifying bifurcations, a technique commonly employed by clinicians~\cite{doi:10.1161/CIRCULATIONAHA.111.032235}. We aim to address this gap by harnessing robot poses and timestamps to reconstruct vessels, thereby incorporating the temporal aspect. By discerning the origin of vascular branches and tracking their progression over time of scan, our method seeks to improve the identification of vessel bifurcations.

\subsubsection{ Real vs.\ Synthetic Data}
\label{realvssynthetic}
Limited availability of US data poses a significant challenge for machine learning applications, especially with limited research focus on the femoral area~\cite{Brattain2018Apr}. When data is scarce, alternative imaging modalities often resort to training on data from physical or digital simulations, which may not generalize well to real-world ultrasound images~\cite{Alkhalifah2022Dec}. Synthetic data may lack crucial realistic features found in reality, such as accurate waveform source signatures, realistic noise, and precise reflectivity, leading to notable discrepancies between datasets~\cite{Alkhalifah2022Dec}.
To address this, we conduct testing on both phantom and real-world data, noting significant differences between them. Phantom data tends to be cleaner, with less densely packed vessels compared to real-world scenarios. Consequently, we use phantom data for experiment development and  validate our findings using real-world data. Fig.~\ref{resus} illustrates the disparity between phantom and real-world experimental data.

%% file: Sections/Method.tex
We combine deep learning techniques with anatomically inspired heuristics to develop an algorithm capable of identifying bifurcations and recommending optimal needle insertion points. An overview can be seen in Figure~\ref{fig:methods}.

\subsection{Vessel Segmentation}
In response to the scarcity of extensive and high-quality US images, we utilize the RESUS algorithm developed by Morales et al.~\cite{Morales2023} for vessel segmentation. RESUS particularly excels in limited datasets such as the one we are working with.

\begin{figure}
\vspace{5pt}
  \centering
  \includegraphics[width=0.80\columnwidth]{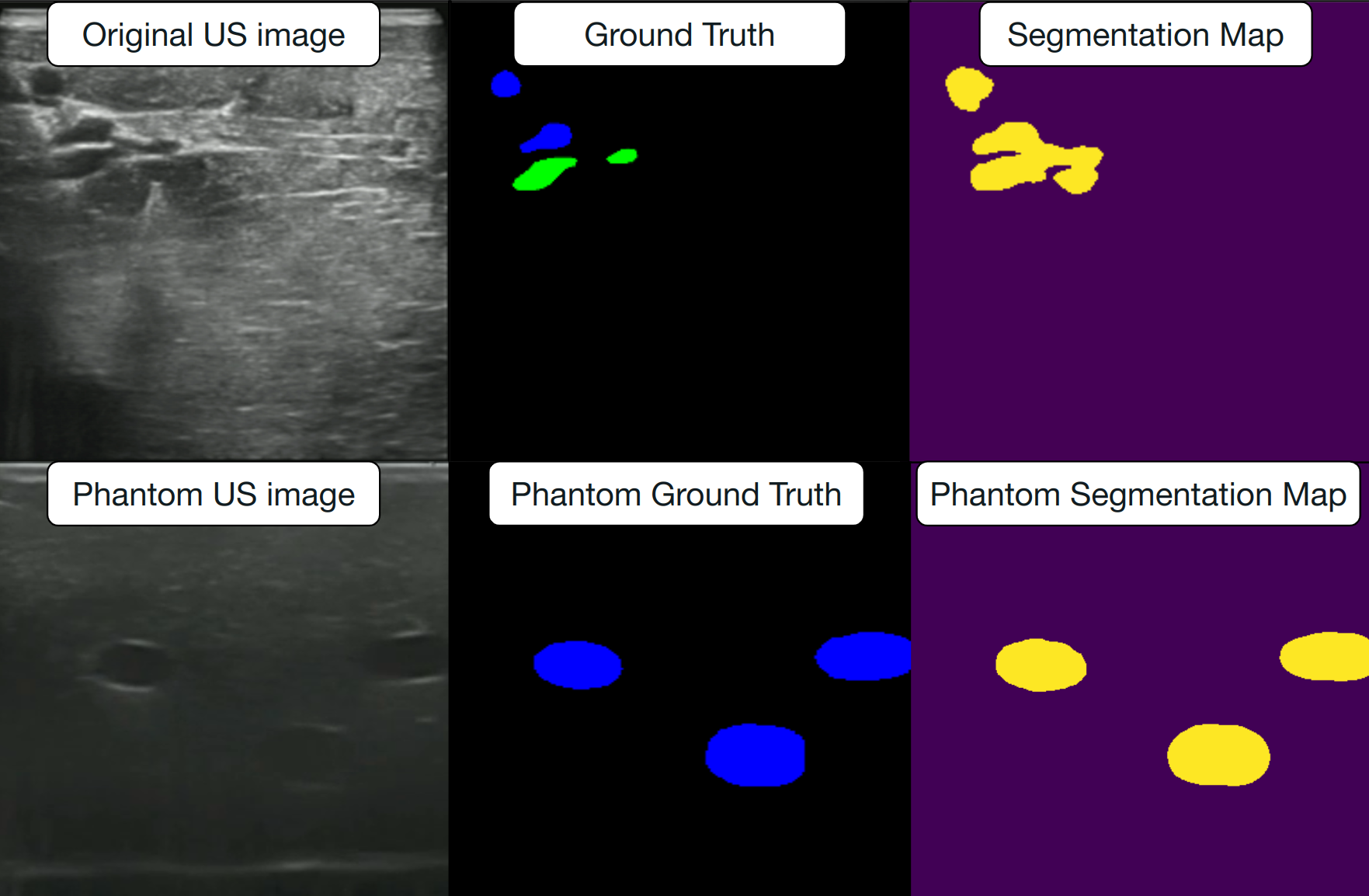}
  \caption{Ultrasound data from pigs and phantoms. The phantom images are cleaner, with vessels appearing well-separated and elliptical. In contrast, the pig images are noisy with asymmetric vessels situated closer together. In the segmentation, these vessels may appear merged, which could be mistaken for a bifurcation if not examined thoroughly.}
   \vspace{-15pt}
  \label{resus}
\end{figure}

Although RESUS yields satisfactory IoU results with our data, it is not without limitations, occasionally exhibiting segmentation inaccuracies and noise artifacts. It also faces a specific challenge when two vessels are placed in close proximity, effectively merging them into a single segmented object, as seen in Fig.~\ref{resus}. This situation poses a substantial risk, especially in the context of bifurcation identification, as it becomes increasingly challenging to differentiate between cases where vessels are adjacent and those where they genuinely bifurcate. If the segmentation process merges these adjacent vessels, crucial differentiation becomes nearly impossible, which motivates the use of spatiotemporal information.

This problem introduces significant risk when considering needle insertion procedures. When attempting to insert a needle between two vessels aligned in parallel, the risk of inadvertently damaging vessel walls escalates, potentially resulting in the formation of hematomas. The occurrence of such hematomas can have dire consequences for patient safety.

\subsection{Centerline Prediction}
\subsubsection{Erosion}
\label{erosion}

To address the issue of adjacent vessels appearing merged, and to remove artifacts and small vessels, we employ an erosion algorithm to post-process the outputs of the segmentation model. The algorithm iteratively removes boundary pixels around the segmented vessels. To do this, we compute the convolution of the segmentation map with a $3\times 3$ kernel with all values 1, scale the result into $[0,1]$, then binarize by thresholding at $0.5$. 
This is repeated until the radius of the minimum enclosing circle is smaller than stopping criterion $\delta_s$ for all segments. Additionally, in scenarios involving distinct vessels, the algorithm aids in their separation, making them easily distinguishable from bifurcations.

\subsubsection{Vessel Detection}


We use the center of the minimum enclosing circle as the position for each segment.
To filter out remaining noise,
we remove any segment where the radius of the circle is less than noise threshold $\delta_{n}$.


\subsubsection{Vessel Tracking}
Vessel tracking involves grouping center points into distinct tracks to represent their spatial configuration. To begin this process, we first transform the vessel centers into temporally arranged 3D point coordinates using the robot pose. Then, we iterate over the timesteps and group these points into tracks to discern individual vessels. Initially, each point in the first frame corresponds to a separate track. Subsequently, new points in each frame are assigned to existing tracks, aiming to minimize the Euclidean distance between the new points and the last point incorporated into the track. Each track can accomodate at most one new point per frame, and this assignment is efficiently computed using the Hungarian Algorithm \cite{Kuhn1955Mar}. A track is terminated if it remains unassigned for five consecutive frames, and if a point cannot be assigned to a new track within a specified threshold distance $\delta_{td}$, a new track is initiated with that point. Tracks with a length of fewer than five center points are discarded. Finally, the tracks are denoised: each track's points are clustered using the Density Spatial Clustering of Applications with Noise (DBSCAN) algorithm~\cite{ester1996clustering}. 
Any outliers identified through this process are subsequently removed from the track.

\begin{table} 
\vspace{5pt}
\centering 
\caption{Empirically determined hyper-parameters for the live pigs and the phantom.} 
\begin{tabular}{>{\centering\arraybackslash\vfill}m{1cm}>{}m{4.3cm}>{\centering\arraybackslash\vfill}m{0.5cm}>{\centering\arraybackslash\vfill}m{1cm}}
\toprule 
\textbf{Variable} & \textbf{Description} &\textbf{Pigs} & \textbf{Phantom}\\ 
\midrule 
$\delta_{n}$ & Threshold for radius of segmented vessel & 3 & 8\\
\midrule
$\delta_{s}$ & Stopping threshold for radius of segmented vessel & 6 & 23 \\ 
\midrule
$\delta_{td}$ & Represents the distance threshold for considering a detected measurement to be part of the same track. & 100 & 100\\
\midrule
$\delta_{h}$ & Maximum difference in average height for tracks to be merged & 200 & 200\\
\midrule
$\delta_{\theta}$ & Minimum angle between best fit lines for tracks to be merged & 10 & 10\\ 
\midrule
$\delta_{sd}$ & Maximum distance for tracks to be merged & 10 & 10\\
\midrule
$\delta_t$ & Minimum threshold to be temporally proximal for bifurcation identification & 0.01 & 0.01 \\
\midrule
$\delta_{bd}$ & Minimum distance to be spatially proximal for bifurcation identification & 10 & 40\\

\bottomrule

\end{tabular}

\label{tab:yourtablelabel} 
\vspace{-9pt}
\end{table}

\subsubsection{Merging Tracks}
We merge the identified tracks based on a set of medical heuristics designed in collaboration with physicians. Our algorithm involves minor adjustments, as shown in Table~\ref{tab:yourtablelabel}, when applied to a medical phantom, as opposed to real-life pigs, due to inherent disparities, discussed in \ref{realvssynthetic}. 

For each track, we compute a least squares line in 3D space that characterizes its center line. The least squares line is 
\begin{equation*}
\textbf{P}_n(\textit{t}_n) = \textbf{S}_n + \textit{t}_n\textbf{V}_n
\end{equation*}
where $\textit{t}_n$ ranges over all real numbers, 
$\textbf{S}_n$ 
is the mean of the points that describe the track, and $\textbf{V}_n$ is the first principal component of differences from the mean, which we obtain by singular value decomposition.  

After that, we use linear interpolation to fill in gaps by adding points to time steps where they are missing in each track. These are used later when identifying bifurcations. 

For each pair of tracks, we check conditions to determine if they can be merged:

\begin{enumerate}
    \item The difference in height (depth from the surface of the skin) of the means is less than $\delta_h$. 
    A single vessel is expected to be at close to the same depth throughout the frame.
    \item The angle between the lines is at least $\delta_\theta$. Otherwise, they are likely to be parallel but distinct vessels.
%
    \item The two tracks intersect within the frame of observation. 
    The intersection can be determined by the minimum distance between the lines, as shown in~\cite{10.5555/2031513}.
        %


        \begin{equation*}
        \begin{aligned}
        \begin{bmatrix}
        t_1^* \\
        t_2^* \\
        \end{bmatrix}
        &= \arg\min_{t_1,t_2} \left\Vert \mathbf{P}_1(t_1) - \mathbf{P}_2(t_2) \right\Vert^2 \\
        &= \frac{1}{(\mathbf{V}_1 \cdot \mathbf{V}_2)^2 - \lVert \mathbf{V}_1\rVert^2\lVert \mathbf{V}_2\rVert^2} \\
        &\quad\times
        \begin{bmatrix}
            - \lVert \mathbf{V}_2\rVert^2 & \mathbf{V}_1 \cdot \mathbf{V}_2 \\
            -\mathbf{V}_1 \cdot \mathbf{V}_2 & \lVert \mathbf{V}_1\rVert^2
        \end{bmatrix}
        \begin{bmatrix}
            (\mathbf{S}_2 - \mathbf{S}_1) \cdot \mathbf{V}_1 \\
            (\mathbf{S}_2 - \mathbf{S}_1) \cdot \mathbf{V}_2
        \end{bmatrix}
        \end{aligned}
        \end{equation*}

    If $(\mathbf{V}_1 \cdot \mathbf{V}_2)^{2} - \lVert\mathbf{V}_1\rVert^{2}\lVert\mathbf{V}_2\rVert^{2}$ $\leq \epsilon_{i}$, where $\epsilon_{i}$ is a numerical tolerance, we consider the lines to be parallel and thus non-intersecting. Otherwise, we find the point of intersection as the midpoint of $\mathbf{P}_1(t^*_1)$ and $\mathbf{P}_2(t^*_2)$. 
    To be considered for merging,
    the point of intersection must be within the bounding box of the union of points from all tracks,
    and the minimum distance between the lines $\left\Vert \mathbf{P}_1(t_1^*) - \mathbf{P}_2(t_2^*) \right\Vert^2$ must be less than a threshold $\delta_{sd}$.

    

\end{enumerate}

To merge tracks, we represent the tracks as nodes on a graph, where two nodes share an edge if the respective tracks meet the conditions for merging. Then the tracks within each connected subgraph are merged into a single track by taking the union of the point sets.

\subsection{Identifying Bifurcations}

Bifurcations are identified through an iterative search within each track, checking if any pair of points meet the following criteria:
\begin{enumerate}
    \item Time between the points is less than $\delta_t$.
    \item Distance between the points is less than $\delta_{bd}$.
    \item The points originate from different tracks before merging.
\end{enumerate}

\subsection{Identifying the Needle Insertion Point}

Given that the robot consistently scans from proximal to distal, the needle insertion spot will be cranially positioned relative to the bifurcation point. Thus we choose the point on the track closest to 2cm away from the bifurcation in the cranial direction. 

%% file: Sections/Experiments.tex
\subsection{Data set}
\begin{figure}
\vspace{5pt}
  \centering
  \includegraphics[width=0.4\columnwidth]{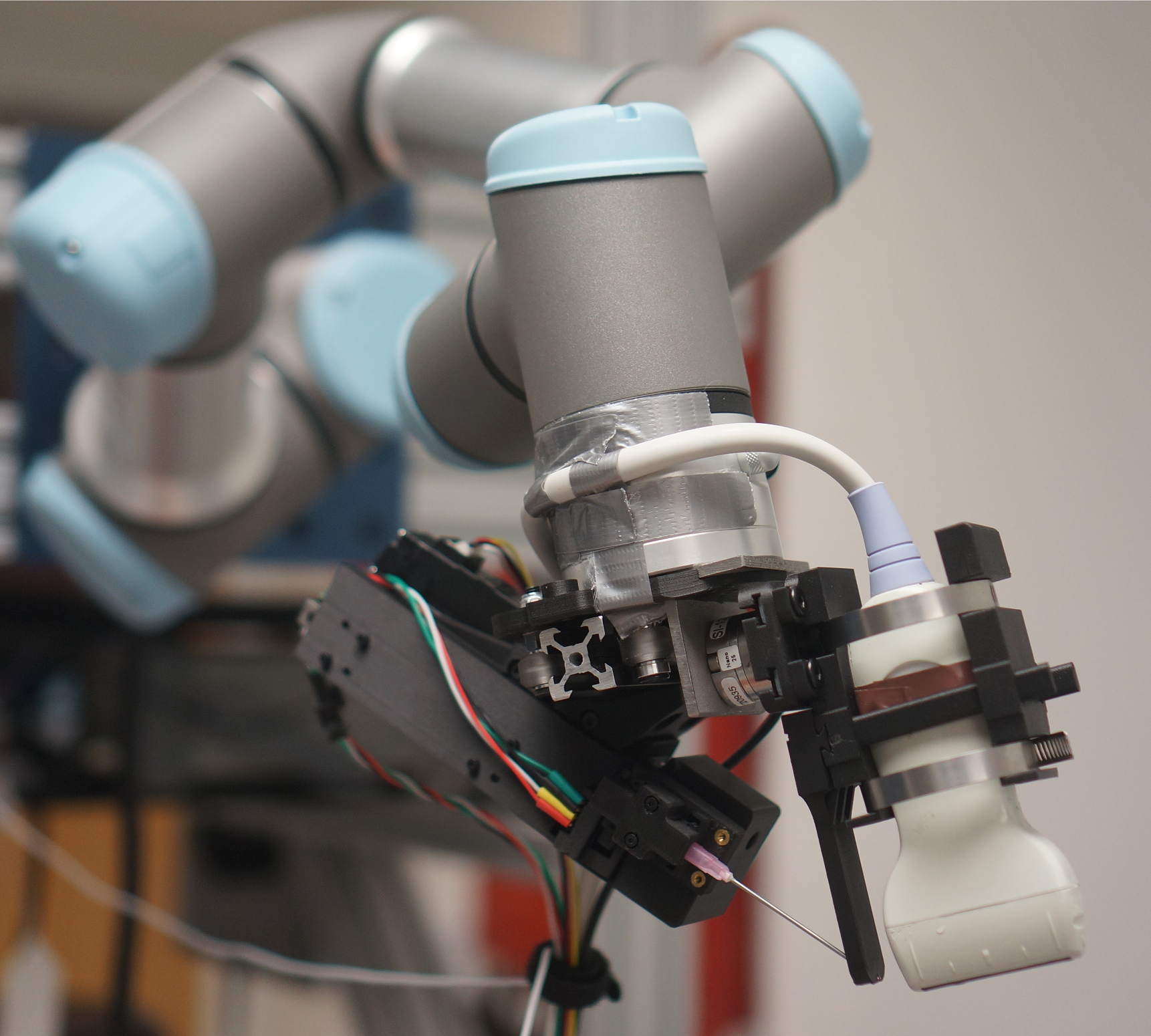}
  \caption{Robotic ultrasound scanning system is shown with a needle insertion mechanism attached to the end-effector of a 6-DOF Universal Robots UR3e Serial Manipulator.}
   \vspace{-15pt}
  \label{fig:robotwithprobe}
\end{figure}

We test BIFURC on both simulated and real-world data. Simulated data was collected from a medical imaging phantom based on the CAE Blue Phantom anthropomorphic gel model\footnote{\url{https://www.caehealthcare.com/solutions/brands/cae-blue-phantom/}}. 
Following successful validation on the phantom, we proceeded to implement our method in a surgical environment at the University of Pittsburgh Medical Center (UPMC) with live pigs under anesthesia. Although stable at the time of collection, the pigs had previously undergone experiments involving hemorrhage and resuscitation, which left them in a weakened state. As a result, their vessels were more closely representative of those encountered in an emergency scenario. These experiments involving live animals were conducted in accordance with the Institutional Animal Care and Use Committee (IACUC) protocol approved by the cognizant authority. We collected 2D ultrasound images of the femoral vessels of six different pigs using a 6-DoF Universal Robot UR3e serial manipulator, as shown in Figure~\ref{fig:robotwithprobe}, to autonomously move the probe used for scanning. The Fukuda Denshi portable point-of-care scanner (POCUS) probe, with a 5MHz linear transducer and a maximum depth of 5cm, was operated at a constant velocity of 0.05m/s, and the US images were recorded at 30 frames per second. An expert surgeon was present during the experiments to ensure the quality of the ultrasound images. Five expert clinicians labeled vessels, bifurcation points, and a needle insertion range using the Computer Vision Annotation Tool (CVAT)~\cite{boris_sekachev_2020_4009388}, ensuring intra-rater reliability in the annotation process. Finally, we used Robot Operating System (ROS)~\cite{ros} to capture time-synchronized robot poses and ultrasound images. \\
The system calibration is conducted once after assembling the robot and does not need to be repeated before each subject. It starts with a preliminary calibration using a robot CAD model to align the robot with the ultrasound system. For refinement, we use a mock phantom with known geometry, containing wires immersed in water, and track these wires with Gaussian fitting to accurately determine their positions in ultrasound images. The robot then follows a predefined trajectory over the wire phantom to calibrate both the time delay between the robot’s movements and the ultrasound measurements, and the transformation between the ultrasound and robot coordinate systems. This process involves computing a transformation matrix to align the ultrasound coordinates with those of the robot and optimizing this matrix to reduce the error between the projected and actual wire positions.

\subsection{Training}
To segment vessels from US images, we used a U-Net architecture with a ResNet34~\cite{7780459} backbone as the encoder, following the methodology outlined in Morales et al.~\cite{Morales2023}. This implementation utilized the Segmentation Models library~\cite{Iakubovskii:2019}.
We train our network on lower resolution ($256 \times 256 $) images, using Dice loss until convergence. We use a batch size of 8, learning rate of $1e-4$ and Adam Optimizer~\cite{DBLP:journals/corr/KingmaB14} for training. These models were trained on a cluster with NVIDIA RTX A6000 GPUs with 48 GiB RAM. We evaluate our models using the leave-one-subject-out cross-validation protocol, in which each pig's data is used once as a test set, while data from the remaining pigs form the training set. RESUS \cite{Morales2023} augmented images are only used for training. To avoid overfitting, the images resliced from the test pig imagery are removed from the training set. 

\subsection{Hyperparameters}
In our experimental setup, we employed a set of hyperparameters to fine-tune the performance of our bifurcation identification and needle insertion algorithm. These hyperparameters govern various aspects of the system, such as the centerline prediction, vessel tracking, merging, and ultimately finding the optimal needle insertion point. It is important to note that these parameters were empirically determined to ensure optimal system performance.

A notable aspect of our experimentation is the distinct hyperparameter configurations of real pigs and our medical phantom. 
While both serve as valuable test subjects, they exhibit distinct characteristics. The medical phantom, designed to mimic human vessels, inherently deviates from the anatomical characteristics of pigs. Consequently, certain hyperparameters required adjustments to effectively accommodate these disparities. 

Despite the limited availability of data, our approach maintained consistent hyperparameters across five pigs, with the exception of one pig flagged as abnormal by clinicians. This consistency highlights the robustness of our algorithm to minor anatomical variations among different pigs. 
Essentially, our parameterization methodology relies on empirical testing and input from clinicians, enabling us to tailor our system's performance to accommodate various subjects, whether they are real pigs or medical phantoms.

%% file: Sections/Results.tex

\begin{table}
\vspace{5pt}
\centering
\caption{Bifurcation Identification Results- All needle insertion spots were inside the optimal range denoted by experts, achieved with single attempts}
\begin{tabular}{>{\centering\arraybackslash\vfill}m{1.3cm} >{\centering\arraybackslash\vfill}m{1cm} >{\centering\arraybackslash\vfill}m{2.3cm} >{\centering\arraybackslash\vfill}m{2.2cm}}
\toprule
\textbf{Subject} & \textbf{IoU Score} & \textbf{Bifurcation Error (mm)}  &  \textbf{Time (secs) } \\
\midrule
Phantom & 0.731 & 5.92 &  1.73 \\
\midrule
Pig 1 & 0.753 & 1.73 &  0.82 \\
Pig 2 & 0.610 & FP \footnotemark, 7.40 &  1.20 \\
Pig 3 & 0.681 & 7.40 & 1.09 \\
Pig 4\footnotemark & 0.610 & 5.01 &  7.26 \\
Pig 5 & 0.693 & 17.72 &  1.99 \\
Pig 6 & 0.606 & 6.68 & 3.30 \\
\midrule
Mean & 0.659 & 7.66 & 2.61  \\
STD & 0.060 & 4.91 &  2.45 \\
\bottomrule
\end{tabular}
\label{tab:bifurcation-accuracy}
\vspace{-10 pt}
\end{table}
\footnotetext[2]{Indicates a second bifurcation was found which was a false positive.}
\footnotetext[3]{Indicates change in hyperparameters $\delta_{t}=0.1$, $\delta_{bd}=20$   }

We evaluate BIFURC using a combination of qualitative and quantitative assessments. Table~\ref{tab:bifurcation-accuracy} presents a summary of our results. The IoU score represents the accuracy of vessel segmentation relative to expert annotations. Bifurcation error quantifies the Euclidean distance between the predicted bifurcation locations within the vessel, and the corresponding ground truth labels provided by expert clinicians. The time column denotes the time it takes our model to identify bifurcations and optimal needle insertion points. Furthermore, we provide the mean and standard deviation of these metrics across the evaluated pigs and phantom. Below, we outline some key observations.\\

\textbf{BIFURC effectively identifies optimal needle insertion points using US images}. We test on six pigs and one phantom. On pigs, BIFURC achieved an 85.7\% success rate in identifying and localizing bifurcations. The error is due to the single false positive; however, even in this instance, the system identified a range that would lead to an optimal needle insertion site. Remarkably, throughout the trials, our robot consistently reported a needle insertion location within the range reported by clinicians. Each identification took an average of 2.61 seconds. Scanning time is not included in this duration. For pig legs, the scanning process typically takes 3-4 seconds. For humans, considering that the common femoral artery is usually about 4 cm long~\cite{swift2023anatomy} and the robot scans at a speed of 0.05 m/s, we estimate the scanning time to be around 2-3 seconds, which aligns with preliminary experimental results.  Additionally, experiments indicate that the robot may require another 1-2 seconds for needle insertion. Therefore, the total time is significantly less than the 185 ± 175 seconds that human experts need for the entire procedure, including both scanning and needle insertion~\cite{Seto}. For context, human experts have an initial success rate of 83\%, but this rate also reflects the overall procedure, including needle insertion. Experts typically make 1.3 attempts per procedure, which affects their success rate. Since BIFURC focuses solely on identifying the optimal insertion points and does not perform needle insertion, it provides a more specific metric of its performance in pinpointing the exact site. BIFURC’s average algorithmic deviation from actual bifurcation points is 7.66 mm, with the optimal insertion site usually 2-5 cm away from the bifurcation~\cite{Bangalore2011}. This indicates that BIFURC is both efficient and accurate in identifying insertion sites, offering significant improvements in time efficiency compared to human experts
and the potential to aid practitioners at various experience levels to achieve near-expert performance.\\



\textbf{Noise in real-world data affects vessel segmentation performance.} A mean IoU of 0.66 across all pigs is a reflection of the noise present in the segmentations. In a qualitative assessment, we reconstruct the vessel centerlines and visually inspect our predicted bifurcation points to ensure they accurately correspond to the bifurcation areas. Although certain instances, such as Pig 5, exhibit larger deviations, we hypothesize that these anomalies may be attributed to segmentation noise, which can occasionally create speckle-like artifacts that influence the algorithm's accuracy. \\

\textbf{Our estimated hyperparameters generalize across most pigs.} Through empirical experiments, we discovered a set of hyperparameters that generalize across the majority of the data, as shown in Table~\ref{tab:yourtablelabel}. We made only slight adjustments to the parameters for bifurcation identification and the kernel erosion algorithm due to the inherent variations in noise levels and spatial arrangements of vessels between phantoms and pigs, as shown in Fig.~\ref{resus}. Also, we made minor adjustments to the bifurcation identification hyperparameters for Pig 4, as physicians noted its vessels displayed highly distinctive shapes, classifying it as abnormal. \\

\textbf{RESUS outperforms other segmentation methods.} To justify our choice of RESUS, we conducted a comparative analysis on our chosen segmentation algorithm against alternative segmentation algorithms used by prior work~\cite{chen2022multiclass, bios11120522}. Specifically, we evaluate segmentation outcomes on several pigs using leave-one-subject out cross-validation. Our results in Table~\ref{tab:resus-vs-yolo} revealed that, while segmentations were not perfect, RESUS consistently outperformed other methods in terms of segmentation quality. 

\begin{table}[ht]
\caption{Segmentation Model Selection}
\begin{tabularx}{\columnwidth}{l*{6}{X}}
\toprule
 & \textbf{RESUS} & \textbf{YOLO (One Class)} & \textbf{YOLO (Multi-Class)} & \textbf{U-Net} & \textbf{3D Bayesian U-Net} & \textbf{2D Bayesian U-Net}\\
\midrule
Pigs A & \textbf{0.70}  & 0.43 & 0.33& 0.64& 0.56 &0.47\\
Pigs B & \textbf{0.67}  & 0.48 &0.20 &0.44 & 0.51 &0.49\\ 
Pigs C & \textbf{0.65}  & 0.49 &0.29 & 0.47 & 0.55&0.52\\ 
Pigs D & \textbf{0.59}  & 0.14 & 0.11&0.15 &0.51 &0.27\\
\midrule
Mean IoU & \textbf{0.65}  & 0.39 & 0.23& 0.43&0.53 &0.44\\
\bottomrule
\end{tabularx}
\label{tab:resus-vs-yolo}
\vspace{-10pt}
\end{table}



%% file: Sections/Conclusion.tex
We leverage domain expertise in developing heuristic-based algorithms integrated with deep learning methods to produce novel results on the problem of vessel bifurcation identification. Our work has several limitations. Firstly, it currently lacks the ability of multi-class segmentation, and the tracking performance is not yet ideal. This is mainly due to the intrinsic cylindrical shape of the vessels, which poses challenges in feature extraction, making differentiation between the arteries and veins a complex task. We recommend using Doppler US and pressure sensing to enable accurate vessel type identification by estimating intravascular pressure. 
Further, our experiments involve anesthetized animals, minimizing subject movement during procedures. We have not studied the potential impact of minor movements on the optimality of the inferred needle insertion points, but in practice, one can perform a final check just before insertion to verify the continued accessibility of the target vessel. 
Lastly, our current in-vivo dataset is modest in size, having a limiting impact on the generalizability and, potentially, effectiveness of our segmentation algorithm. For future work, we plan to increase our subject sample size, explore multiclass segmentation, and study vessels in various anatomical regions to improve the generalizability of the algorithm. Additional factors to consider include tissue deformation during needle insertion, as well as insertion angle and force. They are key to creating a precise and safe intravascular access system. To further assess the reliability and accuracy of the model, we consider incorporating additional validation methods, including CT scans of the same test subjects. Finally, we plan to explore the cognitive burden that the proposed automation may impose on its users.